\documentclass{article}
\usepackage{spconf,amsmath,graphicx}
\usepackage{svg}
\usepackage{hyperref}

\usepackage{enumitem}
\setlist{nosep, leftmargin=14pt}

\usepackage{mwe} 

\title{Improving Acne Image Grading with Label Distribution Smoothing}
\twoauthors
 {Kirill Prokhorov}
	{Openface.io\\
	Vilnius, Lithuania
}
 {Alexandr A. Kalinin\sthanks{Corresponding author: akalinin@broadinstitute.org}\thanks{
 © 2024 IEEE. Personal use of this material is permitted. Permission from IEEE must be
obtained for all other uses, in any current or future media, including
reprinting/republishing this material for advertising or promotional purposes, creating new
collective works, for resale or redistribution to servers or lists, or reuse of any copyrighted
component of this work in other works.
 }}
	{Broad Institute of MIT and Harvard\\
	Cambridge, MA, USA
 }
\begin{document}
%
\maketitle
\begin{abstract}
Acne, a prevalent skin condition, necessitates precise severity assessment for effective treatment.
Acne severity grading typically involves lesion counting and global assessment.
However, manual grading suffers from variability and inefficiency, highlighting the need for automated tools.
Recently, label distribution learning (LDL) was proposed as an effective framework for acne image grading, but its effectiveness is hindered by severity scales that assign varying numbers of lesions to different severity grades.
Addressing these limitations, we proposed to incorporate severity scale information into lesion counting by combining LDL with label smoothing, and to decouple if from global assessment.
A novel weighting scheme in our approach adjusts the degree of label smoothing based on the severity grading scale.
This method helped to effectively manage label uncertainty without compromising class distinctiveness.
Applied to the benchmark ACNE04 dataset, our model demonstrated improved performance in automated acne grading, showcasing its potential in enhancing acne diagnostics.
The source code is publicly available at \url{http://github.com/openface-io/acne-lds}.
\end{abstract}
\section{Introduction}
\label{sec:intro}

\textit{Acne vulgaris}, commonly known as acne, is a widespread skin condition that is estimated to affect over 700 million people worldwide, significantly impacting interpersonal relationships, social functioning, and mental health~\cite{layton2021reviewing}.
Accurate acne severity assessment is important for selecting the right treatment and as a clinical trial outcome~\cite{thiboutot2019assessing}.
However, manual severity grading by visual global assessment and lesion counting is time-consuming and susceptible to inter-observer variability~\cite{lucky1996multirater}.
Moreover, dermatologists are consistently in short supply, particularly in rural areas, and often cases are seen instead by general practitioners with lower diagnostic accuracy, while consultation costs are rising~\cite{resneck2004dermatology,Liu2020}. Therefore, the use of automated tools for computer-aided acne severity assessment may be a promising alternative for broadening the availability of dermatology expertise~\cite{Liu2020}.

Over the last two decades, multiple approaches to automated acne severity assesement from facial photos were proposed.
Initially, these solutions relied on conventional image analysis~\cite{ramli2012acne}, but the advent of deep learning's breakthrough performance improvements in biomedical image analysis have shifted focus to its use for acne lesion detection~\cite{chantharaphaichi2015automatic}, classification~\cite{alamdari2016detection}, counting~\cite{maroni2017automated}, and severity grading~\cite{seite2019development}.

In acne image grading, each photo is assigned a severity level and while over 20 different grading scales have been proposed over time, the medical community has yet to agree on standardized criteria~\cite{agnew2016comprehensive}.
Most grading scales rely on lesion counting as a quantifiable measure informative of severity. 
Recognizing the connection between lesion counting and global severity grading, Wu \textit{et al}.~\cite{Wu2019} introduced a unified framework that tackles both tasks simultaneously and published the benchmark dataset \textit{ACNE04} with annotated lesions.
Utilizing label distribution learning (LDL)~\cite{geng2016label}, their method assigns to each image two label distributions: one for quantifying lesion counts and another for classifying acne severity.
The severity class labels are based on the Hayashi scale~\cite{hayashi2008establishment}, which delineates acne severity into four levels based on lesion count ranges: 1--5 lesions is \textit{mild}, 6--20 lesions is \textit{moderate}, 21--50 lesions is \textit{severe}, and 50+ lesions is \textit{very severe}.

The approach proposed by Wu \textit{et al}.~\cite{Wu2019} generates the ground-truth label distribution for lesion counts independently of the severity scale, assuming the same levels of grade uncertainty for all lesion counts.
However, the predicted lesion count is then converted into the severity grade prediction and evaluated against the labels generated per Hayashi severity scale~\cite{hayashi2008establishment}.
This leads to varying grade uncertainty for different lesion counts based on the severity scale: for example, both 12 and 13 lesion counts confidently correspond to the \textit{moderate} grade, while 20 and 21 lesions are assigned \textit{moderate} and \textit{severe} grades, accordingly.

At the same time, global severity assessment branch directly predicts severity grade distribution from an image.
In contrast with the lesion counting task, accounting for the Hayashi scale is not beneficial here, because the scale delineates severity classes by uneven ranges of lesion counts (e.g., \textit{mild} only includes 1--5 lesions, while \textit{moderate} includes 6--20), making global prediction more challenging.


\begin{figure}[tb]
  \centering
    \includegraphics[clip, trim=0cm 5cm 10cm 0cm, width=8.5cm,]{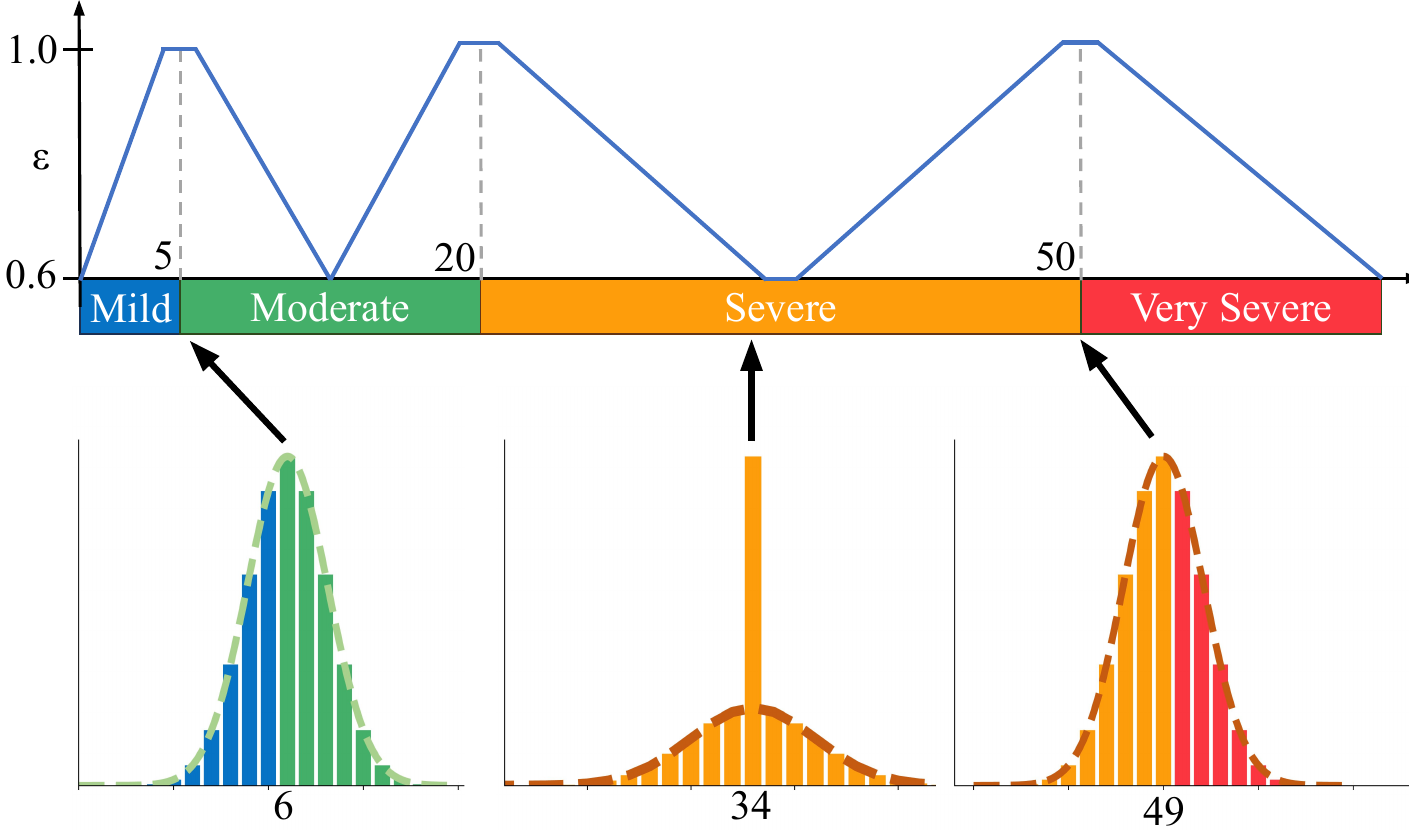}
\caption{Piecewise linear weighting of the smoothing parameter $\varepsilon$ used to control how much of label distribution is added to smooth the hard label. Near the class boundaries $\varepsilon_{min}=1$, which corresponds to LDL, and near the mid-range $\varepsilon=\varepsilon_{min}$, which preserves the dominance of the original label value. The value $\varepsilon_{min}=0.6$ is tuned using single-fold validation.}
\label{fig:epsilon}
\end{figure}

\begin{figure*}[t!]
  \centering{}\includegraphics[width=16.0cm,]{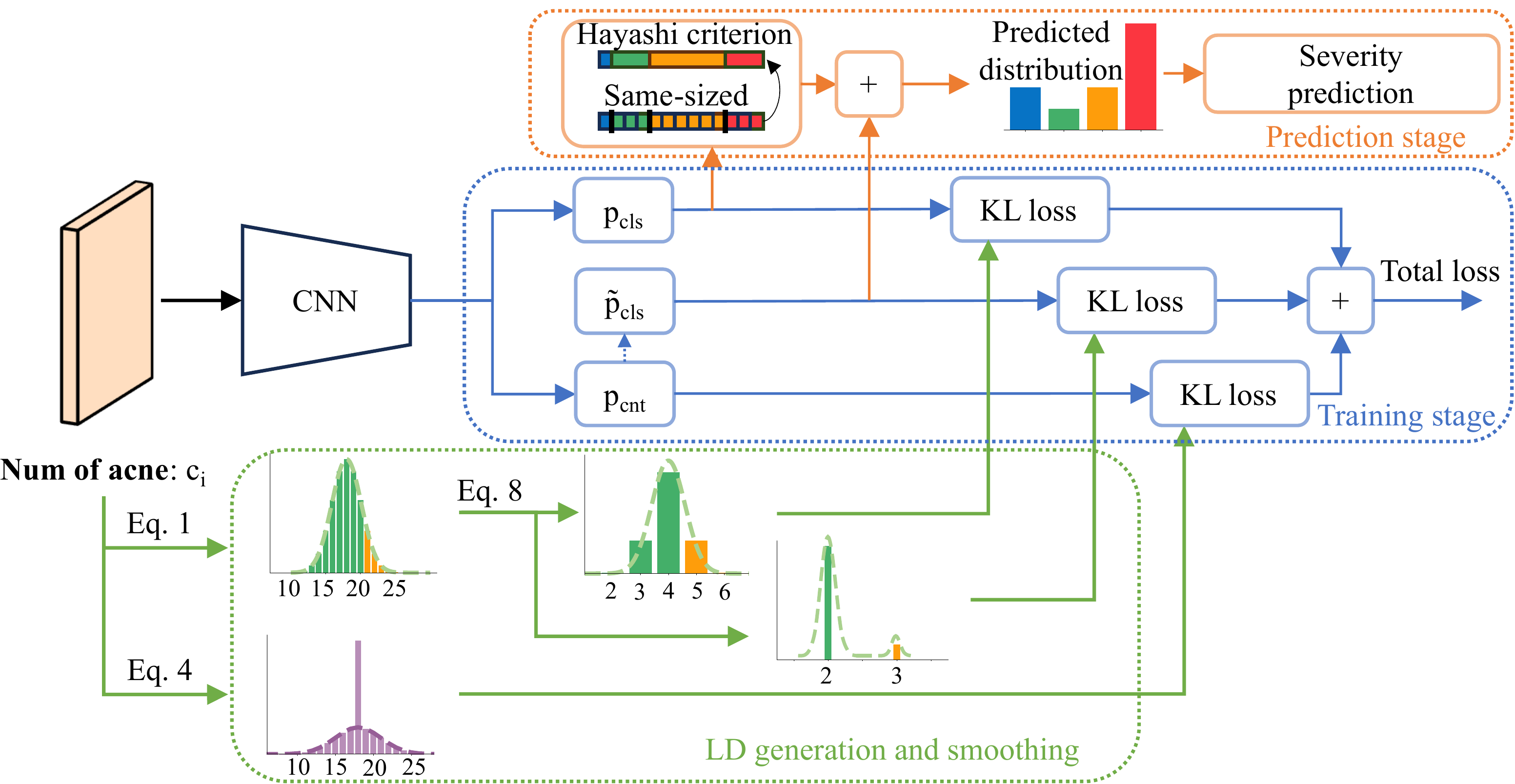}
\caption{Description of the proposed approach. Smoothed label and label distribution are generated using ground-truth count label. After that label distribution is converted into 4 and 13 classes distributions which are used in calculating KL losses along with smoothed distribution. During prediction stage one combines predictions from severity prediction branch with results from counting branch.}
\label{fig:nn}
\end{figure*}

Here, we proposed an approach that addresses these issues by incorporating severity scale information into generating label distributions for lesion counting, while simultaneously removing it from the direct severity grade classification.

Our first contribution can be viewed as a novel way to combine label smoothing~\cite{szegedy2016rethinking} with LDL.
We smooth a hard lesion count label  with the Gaussian label distribution, such that the amount of smoothing depends on where this lesion count falls on the severity scale (Fig.~\ref{fig:epsilon}).
This is realized by introducing a parameter $\varepsilon$ to control amount of smoothing applied to each lesion count based in its proximity to the grade range border.
For the counts at grade range boundaries, we use $\varepsilon=1$ to generate Gaussian label distributions, enabling a soft transition between classes.
This corresponds to LDL, incorporating high grade uncertainty.
But for object counts towards the middle of the grade range, we reduce the weight $\varepsilon$ of the label distribution such that the original count label remains dominant compared to its neighbors.
In such cases, the grade uncertainty is lower, which allows the model to calibrate predictions accordingly.
For instance, an image with a lesion count of 34---well within the range of the \textit{severe} class---generates a distribution with lesser amount of label  smoothing to maintain a highly confident grade prediction (Fig.~\ref{fig:epsilon}).
This hybrid approach ensures that our model accounts for the inherent uncertainty in the counting task without diluting the distinctness of each class.

For the classification branch, we reduce the complexity of the task by breaking down Hayashi-defined uneven grade ranges into evenly-sized classes such that each class range contains exactly five lesion counts.
We demonstrate that our approach improves the results of automated acne grading on the benchmark dataset indicating the potential to improve diagnostics of acne.

\section{Method}
\label{sec:method}

Let \( x_i \) be the \( i \)-th image out of the training set of size \( N \) with the corresponding ground-truth lesion count annotation \( z_i \in \{1, 2, \dots Z\} \), where \( Z \) is the maximum lesion count, and the severity level $y_i\in [1,2,\dots Y]$, where \( Y \) represents the number of distinct severity grades. Overall architecture follows~\cite{Wu2019}, except for changes described below (see Fig.~\ref{fig:nn}).


\subsection{Gaussian label distribution generation}
Wu \textit{et al}.~\cite{Wu2019} used the Gaussian function to generate the lesion count label distribution. For the particular acne count label $c_j$ and image $x_i$ they defined the description degree as:
\begin{equation}
    d_{x_i}^{c_j} = d(c_j|x_i)=\frac{1}{\sqrt{2\pi\sigma^2}M}\exp\left({-\frac{\left(c_j-z_i\right)^2}{2\sigma^2}}\right),
    \label{eq:ldl}
\end{equation}
where \( j \in \{1, 2, \dots, Z\} \) and $M$ is the normalization factor:
\begin{equation}
    M=\frac{1}{\sqrt{2\pi\sigma^2}}\sum\limits_{j=1}^Z \exp\left({-\frac{\left(c_j-z_i\right)^2}{2\sigma^2}}\right),
\end{equation}
such that $d_{x_i}^{c_j} \in \left[0, 1\right]$ and $\sum\limits_{j=1}^Z d_{x_i}^{c_j} = 1$.

\subsection{Label smoothing}
Label smoothing~\cite{szegedy2016rethinking} was proposed to soften the hard
label in the training process to prevent overconfidence and improve generalization. Consider one particular image $x$ with ground-truth label $y_{gt}$ that is one-hot encoded as $q(k|x)=\delta_{k, y_{gt}}$. Then the original label can be replaced with a distribution:

\begin{equation}
    q'(k|x) = (1-\varepsilon)\delta_{k, y_{gt}} + \varepsilon u(k|x),
    \label{eq:smooth}
\end{equation}

\noindent where $u(k|x)$ is usually the uniform distribution $u(k|x)=\frac{1}{K}$, where $K$ is the number of classes. 
As the result, the true label description degree will be reduced, while the other classes will obtain non-zero values. 

\subsection{Scale-adaptive label distribution smoothing}

To obtain more confident predictions for the mid-range counts, while maintaining higher grade uncertainty for counts near the grade border, we propose a methods that combines Gaussian label distribution generation with confident labels via a label smoothing-like weighting scheme (see Fig.~\ref{fig:epsilon}).
We achieve this in two steps.
First, we replace the uniform distribution in eq.~(\ref{eq:smooth}) with the generated label distribution from eq.~(\ref{eq:ldl}).
This limits redistribution of confidence from the hard label to its surrounding neighbors, unlike the traditional label smoothing that assigns some small description degrees to all labels.
Second, we introduce piecewise-linear schedule for the smoothing parameter $\varepsilon$ in order to control the weight of the label distribution base on the count label location in the grading scale, as illustrated on Fig.~\ref{fig:epsilon}.
Now we can replace eq.~(\ref{eq:smooth}) with the following:

\begin{equation}
    q'(c_j|x_i) = [1-\varepsilon(c_j)]q(c_j|x_i) + \varepsilon(c_j) d(c_j|x_i),
    \label{eq:our_smooth}
\end{equation}

\noindent where $q(c_j|x_i)$ is the one-hot encoded ground-truth label, $q'(c_j|x_i)$ is the smoothed label distribution.
Near the class border $\varepsilon=1$, which corresponds to LDL, whereas for the mid-range labels $\varepsilon_{min} \le \varepsilon(c_j) < 1$ ($\varepsilon_{min}$ is a hyperparameter), which is more similar to the traditional label smoothing.

\subsection{Lesion counting branch}
We replace $d_{x_i}^{c_j}$ with $q'(c_j|x_i)$ from eq.~(\ref{eq:our_smooth}) in the loss function that is the Kullback–Leibler (KL) divergence between the generated and predicted distributions eq.~(\ref{fig:nn}):

\begin{equation}
    \mathcal{L}_{cnt}(x_i, z_i) = -\sum\limits_{j=1}^{Z} q'(c_j|x_i) \ln\frac{p_{cnt}(c_j|x_i, \boldsymbol{\theta})}{q'(c_j|x_i)},
    \label{eq:loss_cnt}
\end{equation}

where the probability of image $x_i$ belonging to class $c_j$ is:
\begin{equation}
    p_{cnt}(c_j|x_i, \boldsymbol{\theta})=\exp{(\theta_{c_j})}/\sum\limits_l\exp{(\theta_{l})}.
\end{equation}
Following~\cite{Wu2019}, we also convert count label distributions and their predictions into severity labels and predictions by summing up corresponding probabilities by the Hayashi scale.

\begin{table*}[h]
\centering
\begin{tabular}{|c|c|c|c|c|}

    \hline
	Metric & Wu \textit{et al.}~\cite{Wu2019} & LD smoothing  & New class ranges & Both\\ \hline\hline
    Accuracy     & 83.70 $\pm$ 1.53 & 83.90 $\pm$ 1.48 & 83.63 $\pm$ 1.32 &  \textbf{84.11 $\pm$ 1.94} \\ \hline
	Precision    & 82.97 $\pm$ 1.27 & \textbf{83.38 $\pm$ 3.02} & 82.63 $\pm$ 2.27 & 83.11 $\pm$ 2.56  \\ \hline
	Specificity  & 93.76 $\pm$ 0.63 & 93.81 $\pm$ 0.473 & 93.75 $\pm$ 0.42 & \textbf{ 93.99 $\pm$ 0.68} \\ \hline
	Sensitivity  & 81.06 $\pm$ 3.46 & 81.21 $\pm$ 2.29 & 81.47 $\pm$ 2.88 &  \textbf{81.53 $\pm$ 2.95} \\ \hline
    Youden Index & 74.83 $\pm$ 4.06 & 75.02 $\pm$ 2.75 & 75.22 $\pm$ 3.28 & \textbf{ 75.52 $\pm$ 3.61} \\ \hline
    MCC          & 75.41 $\pm$ 2.35 & 75.69 $\pm$ 2.18 & 75.32 $\pm$ 1.98 & \textbf{ 76.16 $\pm$ 2.82} \\ \hline

\end{tabular}
\caption{Evaluation results on the \textit{ACNE04} dataset~\cite{Wu2019}}
\label{t2}
\end{table*}

\begin{figure}[b!]
  \centering{}\includegraphics[clip, trim=0cm 16cm 26cm 0cm, width=5.5cm,]{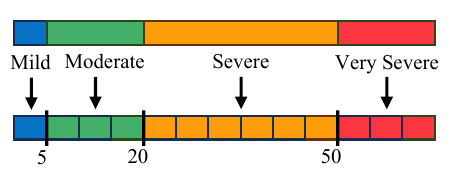}
  \caption{We convert Hayashi scale-based grade ranges into evenly-spaced ones to simplify direct image severity grading.}
  \label{fig:new_cls}
\end{figure}

\subsection{Severity prediction branch}


Since severity grading branch is independent of lesion counting, we can convert Hayashi-based severity grade labels into evenly-spaced ones, see Fig.\ref{fig:new_cls}.
The severity label distribution is generated according to new classes instead of the Hayashi scale.
Then the severity prediction loss function follows:

\begin{equation}
    \mathcal{L}_{cls}(x_i, y_i) = -\sum\limits_{k=1}^{Y'} d(s'_k|x_i) \ln\frac{p_{cls}(s'_k|x_i, \boldsymbol{\theta})}{d(s'_k|x_i)},
    \label{eq:loss_cls}
\end{equation}

\noindent where probability of image $x_i$ to belonging to $s'_k$ class is:

\begin{equation}
    p_{cls}(s'_k|x_i, \boldsymbol{\theta})=\exp{(\theta_{s'_k})}/\sum\limits_l\exp{(\theta_{l})},
\end{equation}

\noindent and $d(s'_k|x_i)$ is the new severity description degree.

\subsection{Combined loss function}
To combine severity grade assessment from the counting branch with direct global grading using the severity prediction branch, we train the model using a multi-task loss function defined as:

\begin{equation}
    \begin{split}
        \mathcal{L}_{i}(x_i, y_i, z_i) = (1-\lambda)\mathcal{L}_{cnt}(x_i,z_i) +\\
        +  \frac{\lambda}{2} \left(\mathcal{L}_{cls}(x_i, y_i) + \mathcal{L}_{cnt2cls}(x_i, y_i)\right),
    \end{split}
    \label{eq:loss_tot}
\end{equation}

\noindent where $\lambda$ is the trade-off hyperparameter.

At the prediction stage, class probabilities $p_{cls}(s'_k|x_i, \boldsymbol{\theta})$ for the new set of classes are converted back to the original Hayashi class probabilities $p_{cls}(s_k|x_i, \boldsymbol{\theta})$ using the reverse mapping, see Fig.~\ref{fig:nn}.
After that, the final predicted distribution is obtained by averaging predicted class and counting probability distributions:

\begin{equation}
p_{tot}(\textbf{s}|x_i, \boldsymbol{\theta})=\frac{1}{2} \left(\tilde{p}_{cls}(\textbf{s}|x_i, \boldsymbol{\theta}) + p_{cls}(\textbf{s}|x_i, \boldsymbol{\theta}) \right).
\end{equation}

\section{Experiments and results}
\label{sec:experiments}

\subsection{Dataset and evaluation details}
\label{subsec:eval_det}

We evaluate the proposed approach using the \textit{ACNE04} benchmarking dataset~\cite{Wu2019}.
It contains $1,457$ images with $18,983$ bounding boxes of lesions.
For evaluating, the dataset is split into 80\% training
set and 20\% testing set, containing $1,165$ and $292$ images, respectively.

Considering accurate acne severity grading as the ultimate goal, we focus on classification metrics to evaluate model performance.
In addition to accuracy, precision, specificity, sensitivity, and Youden Index reported by Wu \textit{et al.}~\cite{Wu2019}, we also added Matthews correlation coefficient (MCC)~\cite{mcc_orig} that has recently been reported to have advantages over other classification metrics~\cite{chicco2020advantages}.
During training, we use maximum validation MCC to select the best epoch for saving the model state for further evaluation.



\subsection{Implementation details}




We were unable to exactly reproduce the results from the original paper by Wu~\textit{et al.}~\cite{Wu2019}.
Therefore, we re-trained their LDL model from scratch using provided source code to ensure fair comparison.
We use exactly the same ResNet-50 \cite{resnet} architecture and training schedule, including the pre-defined $5$-fold cross validation.
We start calculating evaluation metrics after the first learning rate decay event.
We tuned several hyperparameters using a single-fold validation, including the standard deviation $\sigma=3.0$ in eq.~(\ref{eq:ldl}), $\varepsilon_{min} = 0.6$ in eq.~\ref{eq:our_smooth}, and the trade-off parameter $\lambda=0.3$ balancing counting and grading tasks in eq.~(\ref{eq:loss_tot}).



\subsection{Results and ablations}


As shown in Table~\ref{t2}, we compared performance of the baseline approach with both of the proposed contributions and their combination. 
Smoothing labels with generated lesion count label distributions in the scale-adaptive fashion ('LD smoothing' column) immediately demonstrated performance improvement across all metrics.
While the use of evenly-sized class ranges in the severity grading branch showed no obvious improvement when applied independently ('New class ranges' column), the combination of both techniques resulted in further performance boost.
This indicates that the combination of these two components benefits from their complimentary.
The label distribution smoothing method effectively handles the uncertainty at the class boundaries and provides a more nuanced approach to learning the relationship between lesion counts and severity grading, while the simplified class definitions offer a straightforward image grading process for the model.
Together, they balance detail-oriented and global approaches, enhancing overall performance.







\section{Conclusion}
\label{sec:print}

In this work, we introduced an automated acne image grading method that combines smoothing lesion count labels by label distributions based on the severity grading scale and simplifying severity class definitions to enhance global acne grading.
Our results demonstrate the synergy of these strategies, boosting grading accuracy and promising a step forward in automated acne diagnostics.
The novel technique of smoothing hard labels by label distributions instead of the uniform distribution is general and potentially applicable beyond acne grading, for example, for grading tumor malignancy.



\section{Compliance with Ethical Standards}
This research study was conducted retrospectively using human subject data made available in open access~\cite{Wu2019}.


\section{Acknowledgments}
\label{sec:acknowledgments}
The authors thank Natalia Martynova for valuable discussions and other support in development of this project.

\bibliographystyle{IEEEbib}
\bibliography{DS.bib}

\begin{thebibliography}{10}

\bibitem{layton2021reviewing}
AM~Layton, D~Thiboutot, and J~Tan,
\newblock ``Reviewing the global burden of acne: how could we improve care to reduce the burden?,''
\newblock {\em British Journal of Dermatology}, vol. 184, no. 2, pp. 219--225, 2021.

\bibitem{thiboutot2019assessing}
DM~Thiboutot, AM~Layton, M-M Chren, EA~Eady, and J~Tan,
\newblock ``Assessing effectiveness in acne clinical trials: steps towards a core outcome measure set,''
\newblock {\em British Journal of Dermatology}, vol. 181, no. 4, pp. 700--706, 2019.

\bibitem{lucky1996multirater}
Anne~W Lucky, Beth~L Barber, Cynthia~J Girman, Jody Williams, Joan Ratterman, and Joanne Waldstreicher,
\newblock ``A multirater validation study to assess the reliability of acne lesion counting,''
\newblock {\em Journal of the American Academy of Dermatology}, vol. 35, no. 4, pp. 559--565, 1996.

\bibitem{resneck2004dermatology}
Jack Resneck~Jr and Alexa~B Kimball,
\newblock ``The dermatology workforce shortage,''
\newblock {\em Journal of the American Academy of Dermatology}, vol. 50, no. 1, pp. 50--54, 2004.

\bibitem{Liu2020}
Yuan Liu, Ayush Jain, Clara Eng, David~H. Way, Kang Lee, Peggy Bui, Kimberly Kanada, Guilherme {de Oliveira Marinho}, Jessica Gallegos, Sara Gabriele, Vishakha Gupta, Nalini Singh, Vivek Natarajan, Rainer Hofmann-Wellenhof, Greg~S. Corrado, Lily~H. Peng, Dale~R. Webster, Dennis Ai, Susan~J. Huang, Yun Liu, R.~Carter Dunn, and David Coz,
\newblock ``A deep learning system for differential diagnosis of skin diseases,''
\newblock {\em Nature Medicine}, vol. 26, no. 6, pp. 900--908, 2020.

\bibitem{ramli2012acne}
Roshaslinie Ramli, Aamir~Saeed Malik, Ahmad Fadzil~Mohamad Hani, and Adawiyah Jamil,
\newblock ``Acne analysis, grading and computational assessment methods: An overview,''
\newblock {\em Skin Research and Technology}, vol. 18, no. 1, pp. 1--14, 2012.

\bibitem{chantharaphaichi2015automatic}
Thanapha Chantharaphaichi, Bunyarit Uyyanonvara, Chanjira Sinthanayothin, and Akinori Nishihara,
\newblock ``Automatic acne detection for medical treatment,''
\newblock in {\em IC-ICTES}. 2015, IEEE.

\bibitem{alamdari2016detection}
Nasim Alamdari, Kouhyar Tavakolian, Minhal Alhashim, and Reza Fazel-Rezai,
\newblock ``Detection and classification of acne lesions in acne patients: A mobile application,''
\newblock in {\em EIT}. 2016, IEEE.

\bibitem{maroni2017automated}
Gabriele Maroni, Michele Ermidoro, Fabio Previdi, and Glauco Bigini,
\newblock ``Automated detection, extraction and counting of acne lesions for automatic evaluation and tracking of acne severity,''
\newblock in {\em SSCI}. 2017, IEEE.

\bibitem{seite2019development}
Sophie Seit{\'e}, Amir Khammari, Michael Benzaquen, Dominique Moyal, and Brigitte Dr{\'e}no,
\newblock ``Development and accuracy of an artificial intelligence algorithm for acne grading from smartphone photographs,''
\newblock {\em Experimental Dermatology}, vol. 28, no. 11, pp. 1252--1257, 2019.

\bibitem{agnew2016comprehensive}
Tamara Agnew, Gareth Furber, Matthew Leach, and Leonie Segal,
\newblock ``A comprehensive critique and review of published measures of acne severity,''
\newblock {\em The Journal of Clinical and Aesthetic Dermatology}, vol. 9, no. 7, pp. 40--52, 2016.

\bibitem{Wu2019}
Xiaoping Wu, Ni~Wen, Jie Liang, Yu~Kun Lai, Dongyu She, Ming~Ming Cheng, and Jufeng Yang,
\newblock ``Joint acne image grading and counting via label distribution learning,''
\newblock in {\em ICCV}. 2019, pp. 10641--10650, IEEE/CVF.

\bibitem{geng2016label}
Xin Geng,
\newblock ``Label distribution learning,''
\newblock {\em IEEE Transactions on Knowledge and Data Engineering}, vol. 28, no. 7, pp. 1734--1748, 2016.

\bibitem{hayashi2008establishment}
Nobukazu Hayashi, Hirohiko Akamatsu, Makoto Kawashima, and Acne~Study Group,
\newblock ``Establishment of grading criteria for acne severity,''
\newblock {\em The Journal of Dermatology}, vol. 35, no. 5, pp. 255--260, 2008.

\bibitem{szegedy2016rethinking}
Christian Szegedy, Vincent Vanhoucke, Sergey Ioffe, Jon Shlens, and Zbigniew Wojna,
\newblock ``Rethinking the {Inception} architecture for computer vision,''
\newblock in {\em CVPR}. 2016, pp. 2818--2826, IEEE/CVF.

\bibitem{mcc_orig}
Brian~W Matthews,
\newblock ``Comparison of the predicted and observed secondary structure of t4 phage lysozyme,''
\newblock {\em Biochimica et Biophysica Acta (BBA)-Protein Structure}, vol. 405, no. 2, pp. 442--451, 1975.

\bibitem{chicco2020advantages}
Davide Chicco and Giuseppe Jurman,
\newblock ``The advantages of the {Matthews} correlation coefficient ({MCC}) over {F1} score and accuracy in binary classification evaluation,''
\newblock {\em BMC genomics}, vol. 21, no. 1, pp. 1--13, 2020.

\bibitem{resnet}
Kaiming He, Xiangyu Zhang, Shaoqing Ren, and Jian Sun,
\newblock ``Deep residual learning for image recognition,''
\newblock in {\em CVPR}. 2016, pp. 770--778, IEEE/CVF.

\end{thebibliography}

\end{document}